\documentclass{article}

\usepackage{microtype}
\usepackage{graphicx}
\usepackage{subcaption}
\usepackage{booktabs}
\usepackage{multirow}
\usepackage[T1]{fontenc}
\usepackage{listings}
\usepackage{upquote}
\usepackage{tcolorbox}
\tcbuselibrary{listings,breakable,skins}
\usepackage{needspace}

\usepackage{hyperref}

\usepackage[preprint]{icml2026}

\usepackage{amsmath}
\usepackage{amssymb}
\usepackage{mathtools}
\usepackage{amsthm}

\usepackage[capitalize,noabbrev]{cleveref}

\theoremstyle{plain}

\theoremstyle{definition}

\newtheorem{problem}{Problem}

\theoremstyle{remark}

\newcommand{\eg}{\textit{e.g.}}
\newcommand{\ie}{\textit{i.e.}}

\usepackage[textsize=tiny]{todonotes}

\icmltitlerunning{Think Once, Act Fast}

\begin{document}

\twocolumn[

  \icmltitle{Think Once, Act Fast: Cost-Effective Recursive Self-Improvement for Decision Model Harness}

  \icmlsetsymbol{equal}{*}
  \icmlsetsymbol{intern}{*}
  \icmlsetsymbol{corr}{$\dagger$}

  \begin{icmlauthorlist}
    \icmlauthor{Siqi Lai}{hkust,intern}
    \icmlauthor{Haitao Li}{lightspeed,corr}
    \icmlauthor{Hao Liu}{hkust,corr}
  \end{icmlauthorlist}

  \icmlaffiliation{lightspeed}{LIGHTSPEED}
  \icmlaffiliation{hkust}{AI Thrust, Hong Kong University of Science and Technology (Guangzhou)}

  \icmlcorrespondingauthor{Siqi Lai}{slai125@connect.hkust-gz.edu.cn}
  \icmlcorrespondingauthor{Haitao Li}{loganli@lightspeed-studios.com}
  \icmlcorrespondingauthor{Hao Liu}{liuh@ust.hk}

  \icmlkeywords{Machine Learning, ICML}

  \vskip 0.3in
]

\printAffiliationsAndNotice{\textsuperscript{*}This work was done during an internship at LIGHTSPEED. \textsuperscript{$\dagger$}Corresponding authors.}

\begin{abstract}
  Fast System-1 models enable responsive actions but struggle with complex or unfamiliar tasks. Large language model (LLM)-based System-2 provides strong reasoning capabilities, yet repeated invocation incurs substantial latency and computational costs. To bridge this gap, we propose \textbf{JevRSI}, a System-1/System-2 collaboration framework that transforms LLM reasoning into transferable decision support through recursive harness improvement. A Jev-style System-1 model enables low-latency action selection, while an asynchronous System-2 LLM develops a modular harness. The harness integrates procedures, state memory, and world models to provide decision contexts spanning the past, present, and future. Utility-driven retrieval selects modules based on both contextual relevance and its utility on previous tasks. Environmental feedback guides targeted revisions and enables recursive self-improvement. Experiments across three datasets show that JevRSI consistently improves decision quality, matching or surpassing LLM agents in several settings. It preserves low-latency execution with favorable performance–cost trade-offs. Moreover, System-2 development interventions decrease as the harness evolves, amortizing System-2 costs through continuous improvement. Our project is available at \url{https://github.com/usail-hkust/JevRSI}.
\end{abstract}

\section{Introduction}
\label{sec:intro}

Developing agents that act efficiently and accurately across diverse environments remains a fundamental challenge in artificial intelligence. Large language models (LLMs) offer broad knowledge and powerful reasoning capabilities, but their computational overhead and latency hinder time-sensitive applications and tasks (\eg embodied agents, traffic signal control~\citep{lai2025llmlight}). Existing approaches follow three paradigms. System-1 agents employ lightweight models, including rule-based systems, small LLMs, and vision-language-action (VLA) models~\citep{kim2024openvla}. They successfully enable fast decision-making but often struggle with complex or unfamiliar tasks. System-2 agents leverage larger LLMs for deliberate reasoning, offering great flexibility with extensive inference cost~\citep{yao2022react,prasad2023adapt}. System-1/System-2 cooperative methods combine these strengths by selectively invoking sophisticated reasoning while delegating routine decisions to fast executors, aiming to balance decision quality and execution efficiency~\citep{lin2023swiftsage,nvidia2025groot}.

However, existing System-1/System-2 collaboration methods primarily focus on allocating computation by selecting when to invoke System-1 or System-2~\citep{lin2023swiftsage}. Each System-2 invocation still incurs substantially greater latency. This motivates distilling LLM reasoning into reusable support for System-1 while preserving its responsiveness. Jev-style~\citep{almeida2026jev} decision models offer a promising paradigm for this goal. Its structured probabilistic decision interface supports fast, high-quality decision-making and strong generalization, making it well suited to System-1 execution. To equip decision models with LLM reasoning capabilities, we propose recursively improving its harness, constructing informative decision context and guide action selection. LLM-based System-2 leverages interaction feedback to revise the harness, allowing recursive self-improvement to support subsequent decisions. Through this cycle, we aim to continually improve decision quality while preserving the responsiveness of System-1 execution.

Achieving this goal requires addressing two coupled challenges. First, the agent need to accurately identify System-1 failures and determine when to invoke LLM reasoning for harness revision. Delayed intervention may compromise decision quality, while excessive intervention introduces substantial computational overhead. Second, the agent need to leverage real-time environmental feedback to drive continual harness improvement. This requires to accurately identify the underlying harness deficiencies and making targeted improvements to address them. As the harness evolves, the agent continuously diagnoses emerging limitations and refine its decision context accordingly. Together, timely intervention and feedback-driven improvement enable the harness to progressively improve decision quality while preserving the responsiveness of System-1 execution.

To address these challenges, we propose \textbf{JevRSI}, a System-1/System-2 collaboration framework that enables fast execution with recursive harness improvement. 
During execution, a Jev-style decision model serves as System-1, supporting low-latency decisions across tasks and environments. Utility-driven retrieval selects most suitable harness modules to provide informative decision context. 
Meanwhile, System-2 develops three complementary harness modules that provide System-1 with decision contexts spanning the past, present, and future. 
(1) State memory retains exploration history and observed transitions. It is constructed by a task-specific memory-building program for different objectives.
(2) Procedure encodes experience to guide the current action under applicable environmental conditions and task goals.
(3) World model is a predictive model to anticipate future environmental states and action outcomes, enabling counterfactual evaluation of candidate actions.
System-2 uses feedback from ongoing execution to diagnose gaps in these modules and make targeted revisions. Through this recursive development process, JevRSI progressively enhances decision quality and reduces dependence on System-2 with the responsiveness of System-1 execution.

Our contributions are threefold:
(1) We propose JevRSI, the first framework that enables recursive self-improvement of decision-model harnesses during execution, supporting efficient and generalizable decision-making across diverse tasks and environments.
(2) We design a modular harness that integrates exploration history, action guidance, and counterfactual predictions to provide System-1 with informative past, present, and future decision contexts. Feedback-driven harness development further enables continual self-improvement and transfer across execution episodes.
(3) Extensive experiments demonstrate that JevRSI consistently improves decision quality and reduces reliance on System-2 reasoning while preserving real-time responsiveness of System-1 execution. Notably, stronger standalone LLM reasoning capabilities do not necessarily translate into more effective harness development. This finding suggests a promising direction for improving harness development and collaboration mechanisms between LLMs and decision models.
\section{Preliminaries}
\label{sec:prelim}

\begin{problem}[Decision-Model Recursive Self-Improvement]
\label{prob:dm-rsi}
We formulate Decision-Model Recursive Self-Improvement as the continual evolution of an external harness through agent--environment interaction. Let $p_\theta$ be a decision model and $p_\psi$ an harness developer. After $k$ interaction episodes, $H_k$ denotes the current harness.

During episode $k$, the agent executes task $\tau_k$. At step $t$, the harness $M_k$ builds context from the available interaction history $h_{k,t}$, and the decision model selects an action:
\begin{equation}
    a_{k,t}\sim p_\theta\!\left(
        \cdot\mid F_{M_k}(h_{k,t}),\mathcal{A}\right),
    \label{eq:decision-interface}
\end{equation}
where $F_{M_k}$ constructs context from the available interaction history $h_{k,t}$ and $\mathcal{A}$ is the action space. Execution produces evidence $e_k$ (\eg, trajectory, environmental feedback, task outcome). The developer uses this evidence to identify deficiencies and revise the harness:
\begin{equation}
    M_{k+1}=p_\psi(M_k,\tau_k,e_k).
    \label{eq:rsi-update}
\end{equation}
The updated harness guides subsequent actions, whose outcomes inform the next revision. Over a horizon of $B$ episodes, our objective is to recursively refine the harness to maximize the final task success rate while progressively reducing the effort required for harness development:
\begin{equation}
      R_{\mathrm{task}}(M_B)
      = \frac{1}{N}\sum_{i=1}^{N} b_i(M_B) - \sum_{k=0}^{B-1} c_k.
      \label{eq:dm-rsi-objective}
\end{equation}
Here, $b_i(\cdot)\in{0,1}$ indicates success on task $i$, $N$ denotes the number of evaluation tasks, and $c_k$ represents the harness development cost of episode $k$.
\end{problem}


\begin{figure*}[t]
    \centering
    \includegraphics[width=\textwidth]{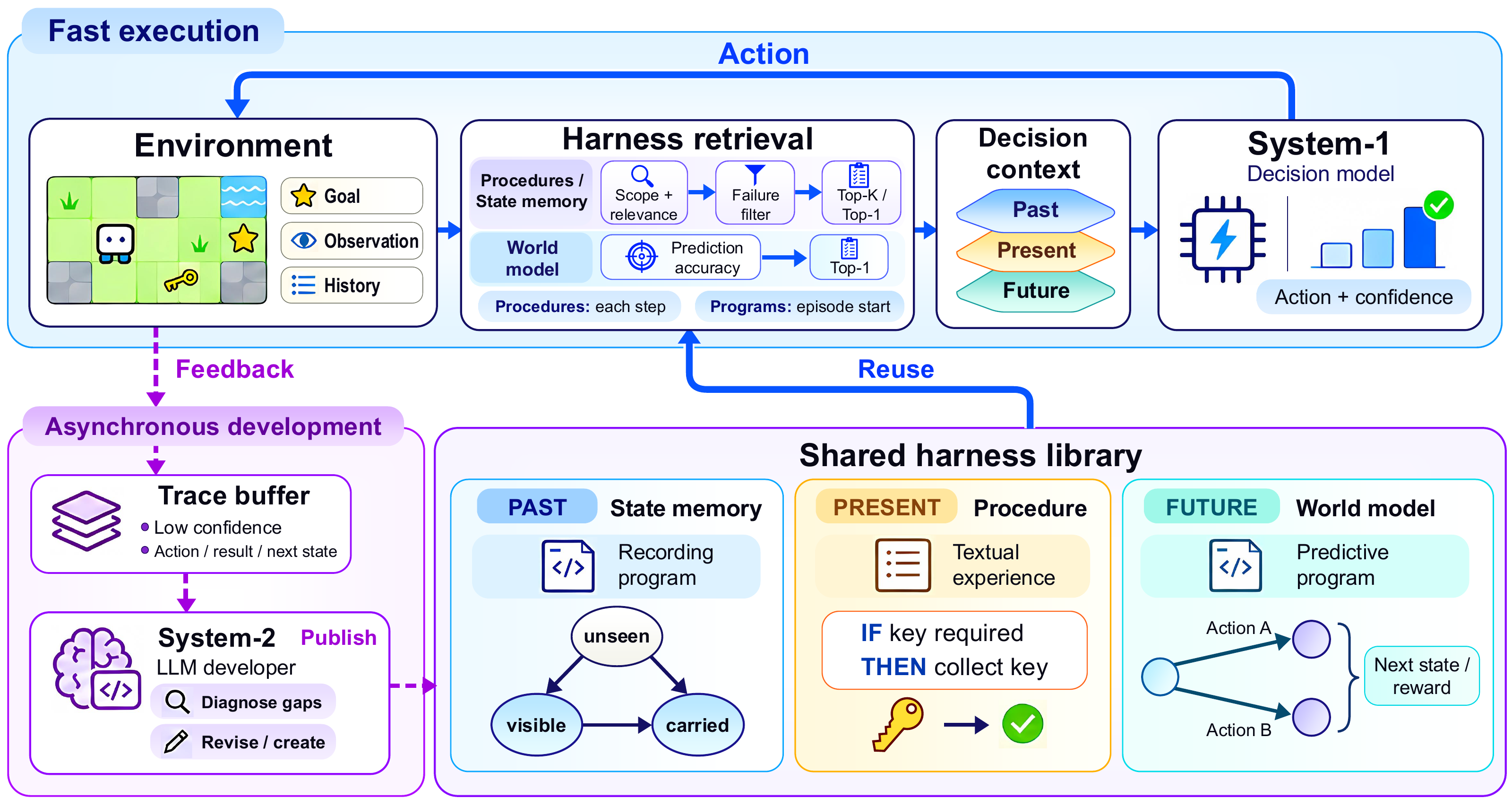}
    \caption{The overview of JevRSI framework.}
    \label{fig:framework}
\vspace{-5pt}
\end{figure*}

\section{Method}
\label{sec:rsi}

\subsection{Dual System Collaboration}
\label{sec:dual-system}


\subsubsection{Real-time Decision Making}
\label{sec:real-time-decision-making}

At step $t$, a query $q_t$ summarizes the task and environmental conditions, including the task goal $g$, observation $o_t$, and candidate actions $\mathcal{A}_t$. Let $\mathcal{M}^t$ denote the latest library of harness modules. We retrieve a set of harness $M_t$ and use them to construct the decision context. 
System-1 selects the action with the highest confidence score based on received context $x_t$, and the runtime executes it in the environment:
\begin{align}
    M_t &= \operatorname{Retrieve}\!\left(
        q_t,\mathcal{M}^t\right), \\
    x_t &= F_{M_t}\!\left(
        q_t, \cdot\right), \\
    a_t &= \arg\max_{a\in\mathcal{A}_t}
        c_\theta(a\mid x_t,\mathcal{A}_t),
\end{align}
where $c_\theta(a\mid x_t,\mathcal{A}_t)$ denotes the confidence score produced by the decision model. After execution, the runtime records the observation, selected modules, action, environmental feedback, and harness. These records provide System-2 with the context to interpret execution outcomes.

\subsubsection{Asynchronous Harness Development}
\label{sec:asynchronous-harness-development}

System-2 develops harness asynchronously while System-1 continues to interact with the environment. We maintain an evidence buffer $\mathcal{B}_t$ to collect interactions associated with low-confidence decisions. After receiving environmental feedback, the runtime updates the buffer according to:
\begin{equation}
    \mathcal{B}_{t+1}=
    \begin{cases}
        \mathcal{B}_t\cup\{e_t\}, & c_t<\eta, \\
        \mathcal{B}_t, & c_t\geq\eta,
    \end{cases}
    \label{eq:development-buffer}
\end{equation}
where $c_t$ denotes the confidence score produced by the decision model, $e_t$ contains the decision context, selected action, observed outcome, and a bounded recent execution trace.

System-2 examines buffered traces with recurring difficulties (\eg repeated failures or stalled task progress). A LLM-based judge reviews the evidence and diagnoses the uncertain decisions. When the potential weakness of current harness modules is identified, System-2 selectively develops or improves a harness module. The newly developed module is published to the shared library $\mathcal{M}^{t+1}$. This design keeps harness development off the real-time decision-making while recursively improving System-1's performance.

\subsection{Decision Model Recursive Self-Improvement}
\label{sec:harness-rsi}


\subsubsection{Decision Model Harness}
\label{sec:modules}

The harness provides three complementary forms of decision support. Procedures supply guidance for the current action. State memory identifies situations that the framework records in exploration graphs. World model predicts future outcomes under alternative actions. Together, they construct past, present, and future context for System-1.

At step $t$, the shared library $\mathcal{M}^t$ contains a finite set of harness modules. Each module is represented by a triple
\begin{equation}
    m=(\kappa_m,\mu_m,b_m),\qquad m\in\mathcal{M}^t,
    \label{eq:harness-module}
\end{equation}
where $\kappa_m\in\{\textsc{proc},\textsc{state},\textsc{wm}\}$ identifies its type (\ie procedure, state memory, or world model), $\mu_m$ contains metadata, and $b_m$ is its body. The metadata records a name, applicable conditions, keywords, and goals. These descriptions form the applicability tag used for retrieval. The metadata also maintains the usage and success counts $n_t(m)$ and $n_t^*(m)$ used to estimate execution utility in \cref{eq:retrieval-utility}. The body contains the module's content, such as the textual guidance and excecutable code, to construct the decision context for System-1.

The retrieval mechanism selects applicable modules $M_t\subseteq\mathcal{M}^t$. Their outputs are assembled into the decision context:
\begin{equation}
    x_t=F_{M_t}\!\left(
        q_t,c_t^{\mathrm{past}},c_t^{\mathrm{now}},c_t^{\mathrm{future}}
    \right),
    \label{eq:temporal-decision-context}
\end{equation}
where the three contextual terms summarize exploration history, current-action guidance, and counterfactual future predictions, respectively.

\paragraph{Procedure}
\label{sec:procedure}

A procedure encodes textual experience distilled by System-2 from earlier interactions. It provides an action recommendation as guidance for the current decision. For a module $m$ with $\kappa_m=\textsc{proc}$, its body is:
\begin{equation}
    b_m=\langle s_m\rightarrow \hat{a}_m\rangle,
    \label{eq:procedure-body}
\end{equation}
where $s_m$ describes the relevant environmental states and task prerequisites, and $\hat{a}_m$ is an action recommendation.

\paragraph{State memory}
\label{sec:state-memory}

For state memory, the body $b_m$ contains a state-memory update program $\sigma_m$ written by System-2. The program specifies a task-dependent state representation and how observed transitions update memory. The framework stores state-action transitions as a directed graph $G_t$:
\begin{equation}
    G_t=(N_t,E_t),\qquad
    E_t\subseteq N_t\times\mathcal{A}\times N_t,
    \label{eq:state-memory-graph}
\end{equation}
Each node is a state representation and edge is a state-action transition. After executing $a_t$, the runtime compacts the resulting observation into a new state representation $v_{t+1}$. It then updates the state-action transition graph:
\begin{equation}
\begin{aligned}
    N_{t+1}&=N_t\cup\{v_{t+1}\}, \\
    E_{t+1}&=E_t\cup\{(v_t,a_t,v_{t+1})\}, \\
    G_{t+1}&=(N_{t+1},E_{t+1}).
\end{aligned}
\label{eq:state-graph-context}
\end{equation}
The framework tracks the visit count and most recent visit step for each node, as well as the traversal count for each edge. During System-1 decision making, the runtime extracts a compact summary from the graph. Summaries from retrieved state-memory modules form $c_t^{\mathrm{past}}$, providing System-1 with relevant exploration history and observed state transitions. This context helps System-1 recognize previously visited states and avoid redundant exploration.

\paragraph{World model}
\label{sec:world-model}

For world model, the body $b_m$ contains a one-step action-state prediction program $\omega_m$ written by System-2. The program predicts the possible next states after taking different actions:
\begin{align}
    \widehat{Y}_{t+1} &= \omega_m(s_t,G_t,\mathcal{A}_t), \\
    \widehat{Y}_{t+1} &= {(s_{t+1}^a, R(s_t|a), \textsc{DONE}^{(s_t|a)}_{t+1})_{a\in\mathcal{A}_t}},
\end{align}
where $s_{t+1}^a$ is the predicted next state after taking action $a$, $R(s_t|a)$ is the predicted reward after taking action $a$, and $\textsc{DONE}^{(s_t|a)}_{t+1}$ is a boolean indicating whether the task is done after taking action $a$. This context enables System-1 to compare anticipated action effects before acting.

\subsubsection{Utility-Driven Harness Retrieval}
\label{sec:utility-driven-harness-retrival}

Our retrieval mechanism combines contextual relevance with utility constraints derived from execution feedback. It retrieves three types of harness modules with distinct strategies. State-memory programs are selected at the beginning of each episode to maintain exploration history. Procedures are retrieved at each decision step to guide current actions. World models are selected based on their predictive accuracy of action consequences. Together, these modules provide System-1 with past, present, and future decision contexts.

For procedures and state memory modules, retrieval considers both contextual relevance and execution history. The query $q_t$ includes the task goal, current observation and interaction context. Each module is indexed by its description and applicability metadata. Let $\widehat{M}_t \subseteq \mathcal{M}^t$ denote the modules whose declared scope matches the current context. We compute relevance scores and select candidates as:
\begin{align}
&s_t(m) = \operatorname{Rel}(q_t,\operatorname{tag}(m)), \\
&\mathcal{C}_t = {m\in \widehat{M}_t : s_t(m)>0,\ f_t(m)<L},
\end{align}
where $\operatorname{Rel}$ measures textual relevance, $f_t(m)$ counts consecutive failed episodes associated with the module $m$, and $L$ is the failure limit. Modules exceeding this limit remain excluded until revised by System-2.

We retrieve $K$ most relevant procedures at each step:
\begin{equation}
M_t^{\mathrm{proc}}=\operatorname{TopK}_{m\in\mathcal{C}_t^{\mathrm{proc}}}s_t(m),
\end{equation}
where $\mathcal{C}_t^{\mathrm{proc}}$ denotes the eligible procedure subset. For state memory, the runtime selects the most relevant program at the beginning of each episode. World models follow a separate selection criterion based on prediction accuracy. The selected modules collectively form $M_t$, providing System-1 with compact decision support.

\subsubsection{Recursive Harness Improvement}
\label{sec:recursive-harness-improvement}

Recursive harness improvement is driven by feedback from environmental interactions. After System-1 executes $a_t$, the environment transitions to $s_{t+1}$. These reveal action outcomes and observed state changes. The runtime records:
\begin{equation}
    e_t^{\mathrm{env}}=(M_t, U_t, a_t,R(s_t|a_t),s_{t+1}),
    \label{eq:environment-feedback-record}
\end{equation}
where $U_t = \{u_t(m)\}_{m\in M_t}$ is the utility of the harness modules used during execution. The runtime attributes each episode's outcome to the modules used during execution. Their utility score is defined as:
\begin{equation}
    u_t(m)=
    \begin{cases}
        n_t^*(m)/n_t(m), & n_t(m)>0, \\
        0, & n_t(m)=0,
    \end{cases}
    \label{eq:retrieval-utility}
\end{equation}
where $n_t(m)$ and $n_t^*(m)$ denote the numbers of executed episodes and successful episodes, respectively. This utility informs subsequent System-2 development and revisions.

The runtime combines interaction feedback with decision context to construct evidence records $e_t$. Low-confidence decisions are added to the development buffer $\mathcal{B}_t$, triggering reviews of procedures and state-memory modules. Separately, world-model development is triggered after an episode when an executed future state contradicts a model prediction. Based on the collected feedback, System-2 diagnoses incorrect guidance, inaccurate predictions, and missing module coverage. It then proposes targeted updates:
\begin{equation}
    (M^{\text{new}}_t,M^{\text{rev}}_t)
        =p_\psi(\mathcal{M}^t,\mathcal{B}_t),
    \label{eq:recursive-harness-proposal}
\end{equation}
where $M_t^{\mathrm{new}}$ and $M_t^{\mathrm{rev}}$ is newly proposed and revised modules, respectively. Revisions address observed deficiencies, while new modules extend coverage to previously unsupported environmental conditions or task requirements. The resulting modules are incorporated into the shared library $\mathcal{M}^{t+1}$ and become available for retrieval in subsequent decisions. This process establishes a recursive improvement loop that continuously guides harness development.

\vspace{-5pt}
\section{Experiments}
\label{sec:exp}


\subsection{Experiment Settings}
\label{sec:exp-settings}

\paragraph{Benchmarks and splits.}
We evaluate JevRSI on ALFWorld~\citep{shridhar2020alfworld}, BabyAI~\citep{chevalierboisvert2018babyai}, and TextCraft~\citep{prasad2023adapt}. Each benchmark is divided into three disjoint splits: training, in-distribution evaluation (ID), and out-of-distribution evaluation (OOD). The training split supplies interaction trajectories for initial harness development. ID tasks assess performance on unseen instances from the training distribution. OOD tasks assess generalization under distribution shifts. The statistics of the datasets are shown in \cref{tab:dataset-stats}.

\begin{table}[t]
    \centering
    \setlength{\tabcolsep}{15pt}
    \caption{Statistics of the datasets.}
    \label{tab:dataset-stats}
    \resizebox{\columnwidth}{!}{
    \begin{tabular}{lccc}
        \toprule
        & \multicolumn{1}{c}{ALFWorld} & \multicolumn{1}{c}{BabyAI} & \multicolumn{1}{c}{TextCraft} \\
        \midrule
        Training size & 200 & 200 & 200 \\
        ID size & 140 & 150 & 150 \\
        OOD size & 134 & 150 & 150 \\
        \bottomrule
    \end{tabular}}
\vspace{-15pt}
\end{table}

\vspace{-8pt}
\paragraph{Comparison methods.}
We compare JevRSI against three groups: vanilla decision models, system-1/2 cooperative agents, and LLM agents. For vanilla decision models, we directly use the decision models with environment interaction histories for evaluation. For system-1/2 cooperative agents, we use JevHarness~\citep{tianyucodings2026jevharness} and JevAgentHarness~\citep{zhu2026jevagentharness} as the baselines. JevHarness employs an LLM agent to develop the harness before deployment, which remains fixed throughout evaluation. JevAgentHarness cooperates a decision model with an LLM agent in a cost-effective way, invoking the LLM to take over decision-making when the decision model's confidence falls below 0.8. For LLM agents, we adopt the same framework as JevRSI but replace the System-1 with the evaluated LLM.

\vspace{-8pt}
\paragraph{Backbone models.}
The evaluated LLMs are GPT-6.1 Sol~\citep{openai2026gpt61sol}, DeepSeek-V4.1-Flash~\citep{deepseek2026v41flash}, and Qwen3.8-27B~\citep{qwen2026qwen38}. The decision models are Jev-1.14~\citep{typesafe2026jev114} and StartLux (9B and 27B)~\citep{startlux2026decision}. We uniformly set JevHarness and JevAgentHarness to use DeepSeek-V4.1-Flash and StartLux-27B as the LLM and decision model, respectively. Task observation, action access, and interaction limits are held consistent within each dataset.

\begin{table}[t]
      \centering
      \setlength{\tabcolsep}{2pt}
      \caption{Overall comparison results. We use Success Rate@1 (SR@1) to measure the performance of agents. Parentheses indicate relative improvement over the vanilla decision model.}
      \label{tab:overall-comparison}
      \footnotesize
      \newcommand{\gain}[2]{\begin{tabular}{@{}c@{}}#1\\[-2pt]{\scriptsize(#2)}\end{tabular}}
      \resizebox{\columnwidth}{!}{
      \begin{tabular}{l|cc|cc|cc}
          \toprule
          \multirow{3}{*}{\textbf{Baselines}} & \multicolumn{2}{c}{ALFWorld} & \multicolumn{2}{c}{BabyAI} & \multicolumn{2}{c}{TextCraft} \\
          \cmidrule(lr){2-3} \cmidrule(lr){4-5} \cmidrule(lr){6-7}
          & ID & OOD & ID & OOD & ID & OOD \\
          \midrule
          \multicolumn{7}{c}{LLM Agents} \\
          \midrule
          GPT-6.1 Sol & \textbf{93.57\%} & \textbf{96.26\%} & \textbf{99.33\%} & \textbf{100.0\%} & \textbf{100.0\%} & \textbf{82.00\%} \\
          DeepSeek-V4.1-Flash & 80.71\% & 82.83\% & 72.66\% & \underline{92.00\%} & \textbf{100.0\%} & 62.00\% \\
          Qwen3.8-27B & 82.85\% & 76.11\% & \underline{78.66\%} & 64.00\% & \textbf{100.0\%} & 55.33\% \\
          \midrule
          \multicolumn{7}{c}{Decision Models} \\
          \midrule
          Jev & 46.42\% & 36.56\% & 28.66\% & 34.00\% & 94.66\% & 22.00\% \\
          StartLux-9B & 46.42\% & 44.77\% & 22.00\% & 25.33\% & 86.00\% & 4.66\% \\
          StartLux-27B & 45.71\% & 45.52\% & 36.00\% & 40.00\% & \textbf{100.0\%} & 62.66\% \\
          \midrule
          \multicolumn{7}{c}{System-1/2 Cooperative Agents} \\
          \midrule
          JevHarness & 60.71\% & 59.70\% & 20.00\% & 21.33\% & 52.66\% & 1.33\% \\
          JevAgentHarness & 68.57\% & 81.34\% & 58.00\% & 50.00\% & 86.00\% & 4.00\% \\
          \midrule
          \multicolumn{7}{c}{JevRSI (Decision-model / LLM)} \\
          \midrule
          Jev / GPT & \gain{52.14\%}{+12.32\%} & \gain{39.55\%}{+8.18\%} & \gain{26.00\%}{$-$9.28\%} & \gain{37.33\%}{+9.79\%} & \gain{\underline{96.66\%}}{+2.11\%} & \gain{20.66\%}{$-$6.09\%} \\[1.5pt]
          \midrule
          Jev / DeepSeek & \gain{62.14\%}{+33.86\%} & \gain{61.19\%}{+67.37\%} & \gain{39.33\%}{+37.23\%} & \gain{40.00\%}{+17.65\%} & \gain{97.33\%}{+2.82\%} & \gain{23.33\%}{+6.05\%} \\[1.5pt]
          \midrule
          Jev / Qwen & \gain{63.57\%}{+36.95\%} & \gain{52.23\%}{+42.86\%} & \gain{31.33\%}{+9.32\%} & \gain{38.00\%}{+11.76\%} & \gain{94.66\%}{0.00\%} & \gain{20.00\%}{$-$9.09\%} \\[1.5pt]
          \midrule
          StartLux-9B / GPT & \gain{55.71\%}{+20.01\%} & \gain{52.98\%}{+18.34\%} & \gain{23.33\%}{+6.05\%} & \gain{24.00\%}{$-$5.25\%} & \gain{86.66\%}{+0.77\%} & \gain{6.00\%}{+28.76\%} \\[1.5pt]
          \midrule
          StartLux-9B / DeepSeek & \gain{56.42\%}{+21.54\%} & \gain{51.49\%}{+15.01\%} & \gain{20.00\%}{$-$9.09\%} & \gain{23.33\%}{$-$7.90\%} & \gain{88.66\%}{+3.09\%} & \gain{4.00\%}{$-$14.16\%} \\[1.5pt]
          \midrule
          StartLux-9B / Qwen & \gain{73.57\%}{+58.49\%} & \gain{76.11\%}{+70.00\%} & \gain{27.33\%}{+24.23\%} & \gain{27.33\%}{+7.90\%} & \gain{94.66\%}{+10.07\%} & \gain{6.00\%}{+28.76\%} \\[1.5pt]
          \midrule
          StartLux-27B / GPT & \gain{55.00\%}{+20.32\%} & \gain{58.20\%}{+27.86\%} & \gain{50.66\%}{+40.72\%} & \gain{55.33\%}{+38.33\%} & \gain{\textbf{100.0\%}}{0.00\%} & \gain{70.00\%}{+11.71\%} \\[1.5pt]
          \midrule
          StartLux-27B / DeepSeek & \gain{\underline{86.42\%}}{+89.06\%} & \gain{\underline{85.82\%}}{+88.53\%} & \gain{65.33\%}{+81.47\%} & \gain{68.66\%}{+71.65\%} & \gain{\textbf{100.0\%}}{0.00\%} & \gain{64.00\%}{+2.14\%} \\[1.5pt]
          \midrule
          StartLux-27B / Qwen & \gain{84.28\%}{+84.38\%} & \gain{82.83\%}{+81.96\%} & \gain{64.00\%}{+77.78\%} & \gain{64.00\%}{+60.00\%} & \gain{\textbf{100.0\%}}{0.00\%} & \gain{\underline{70.66\%}}{+12.77\%} \\
          \bottomrule
      \end{tabular}}
  \vspace{-15pt}
  \end{table}

\vspace{-5pt}
\subsection{Overall Comparison}
\label{sec:exp-overall}

\cref{tab:overall-comparison} presents the overall results across three datasets. Vanilla decision models often achieve success rates below 50\%, and larger models generally perform better. LLM agents achieve higher success rates, with GPT-6.1 Sol exceeding 90\% in most settings. These results highlight the benefits of LLM reasoning for long-horizon interactions. However, existing System-1/System-2 cooperative frameworks show inconsistent improvements. JevHarness improves ALFWorld but underperforms vanilla models on BabyAI and TextCraft, suggesting limitations of static harness without recursive improvement in dynamic environments and unseen tasks. JevAgentHarness achieves 81.34\% and 50.00\% OOD success on ALFWorld and BabyAI, respectively, but only 4.00\% on TextCraft. This suggests that confidence-based LLM intervention may fail to prevent harmful decisions made by System-1 in challenging tasks.

In contrast, JevRSI consistently improves decision-model performance, matching or surpassing LLM agents in several settings. With Qwen as System-2, StartLux-9B improves from 44.77\% to 76.11\% on ALFWorld OOD, matching the Qwen LLM agent. On TextCraft OOD, StartLux-27B achieves 70.66\%, outperforming both its vanilla counterpart (62.66\%) and the Qwen agent (55.33\%). Notably, strong LLM-agent performance does not necessarily translate into effective harness development. Although GPT-6.1 Sol achieves the highest standalone performance, it provides weaker harness support in some settings. On BabyAI with StartLux-27B, GPT initiates only 10 development steps and achieves lower success rates. Compared with Qwen and DeepSeek, they initiate 67 and 49 steps, respectively. These findings highlight that effective System-2 development requires both strong reasoning and accurate identification of opportunities for harness improvement.

\begin{figure}[t]
    \centering
    \includegraphics[width=\columnwidth]{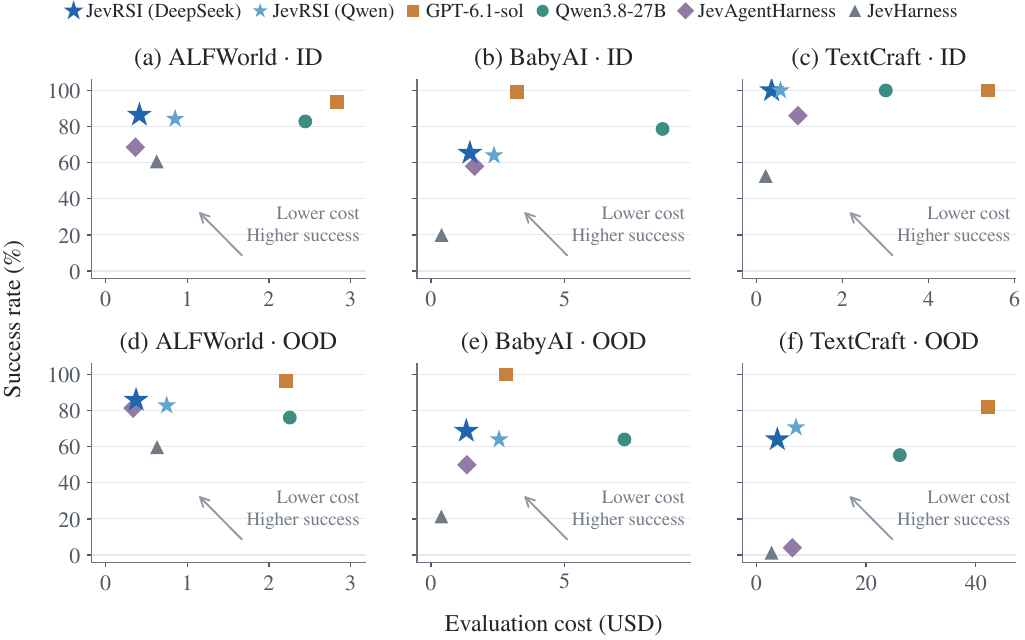}
    \caption{Cost-effectiveness comparison results. All decision-model calls are uniformly priced using the TypeSafe Jev API.}
    \label{fig:cost-effectiveness}
\vspace{-10pt}
\end{figure}

\subsection{Cost-Effectiveness Comparison}
\label{sec:exp-cost}

\begin{figure}[t]
    \centering
    \includegraphics[width=\columnwidth]{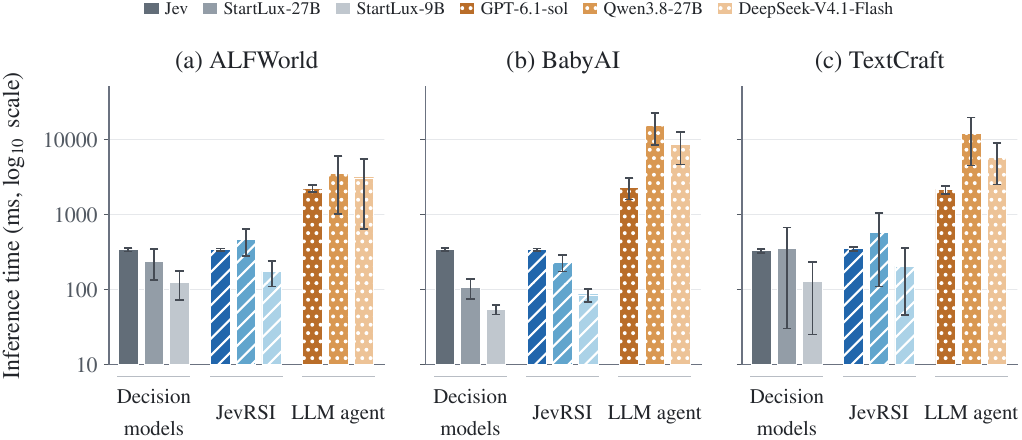}
    \caption{Inference time comparison results.}
    \label{fig:inference-time}
\end{figure}

\begin{figure}[t]
    \centering
    \includegraphics[width=\columnwidth]{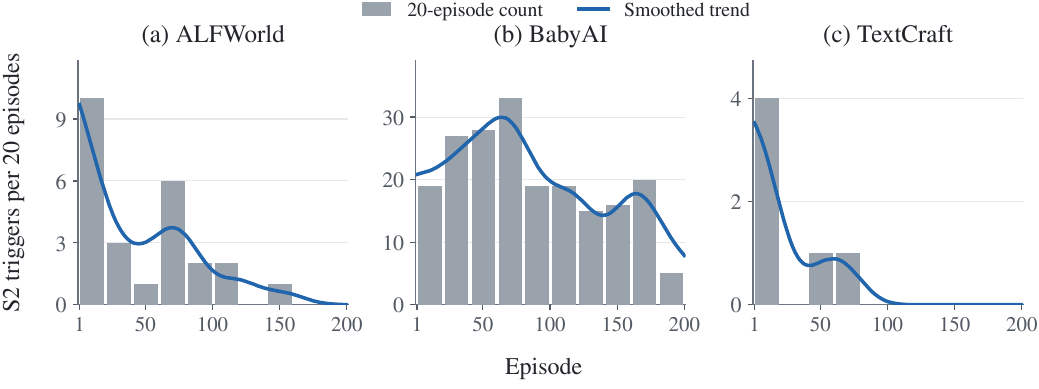}
    \caption{Harness development step distribution in training split.}
    \label{fig:harness-development-step-distribution}
\vspace{-10pt}
\end{figure}

We evaluate cost-effectiveness along three dimensions: token-based expenditure, decision-model inference time, and System-2 development frequency. The \cref{fig:cost-effectiveness,fig:inference-time,fig:harness-development-step-distribution} show that JevRSI achieves strong task performance with lower inference costs and requires fewer development interventions as the harness modules are gradually improved.

\cref{fig:cost-effectiveness} shows the success rate and token-based expenditure of JevRSI and LLM agents during two evaluation splits. It demonstrates a favorable trade-off between task success and token-based expenditure. For example, JevRSI with DeepSeek achieves the same 100\% success rate as the GPT agent at approximately \$0.36 versus \$5.38, reducing evaluation cost by about 93\% on TextCraft ID. This result shows that reusable harness modules preserve task performance while reducing expenditure on LLM reasoning.

\cref{fig:inference-time} illustrates the inference time of vanilla decision models, JevRSI and LLM agents during two evaluation splits. It shows that JevRSI preserves the low inference regime as vanilla decision models. Its mean System-1 inference time ranges from 0.08 to 0.58 seconds across the three benchmarks. LLM agents require 2.14--15.58 seconds on average. Thus, harness-supported action selection retains fast model responses while improving decision quality. Asynchronous development allows System-1 to implement the harness modules without waiting for System-2 development. These results demonstrate that JevRSI achieves real-time decision-making while maintaining high task performance.

\cref{fig:harness-development-step-distribution} shows the number of System-2 harness development steps in the training split. It shows that System-2 development becomes less frequent as the harness modules are gradually improved. Between the first and second halves of training, System-2 triggers fall from 22 to 3 on ALFWorld and from 126 to 75 on BabyAI. TextCraft requires six triggers in the first half and none in the second. Although activity fluctuates locally, later episodes require fewer development interventions overall. This trend suggests that reusable modules reduce recurring development needs and create opportunities to amortize System-2 costs over continued interaction after recursive harness improvement.

\begin{figure}[t]
    \centering
    \includegraphics[width=\columnwidth]{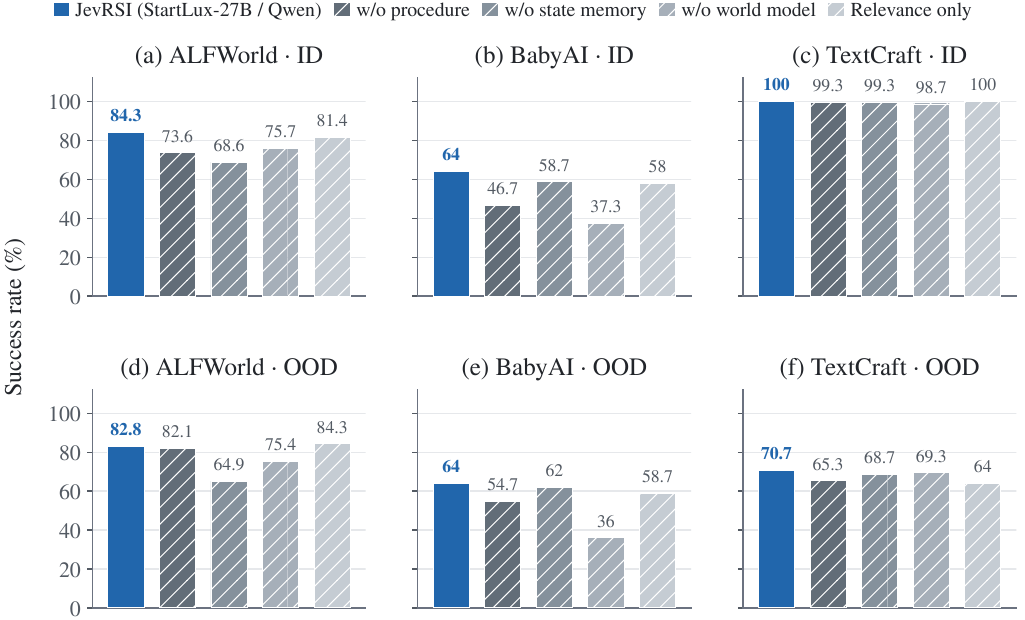}
    \caption{Ablation study results over four JevRSI variants.}
    \label{fig:ablation-studies}
\vspace{-10pt}
\end{figure}

\subsection{Ablation Studies}
\label{app:ablation}

\cref{fig:ablation-studies} evaluates the contributions of each harness module and utility-driven retrieval using StartLux-27B and Qwen. Removing any module reduces success across all six settings, but their impact varies by task. State memory contributes most on ALFWorld, where its removal reduces ID and OOD success by 15.7 and 17.9 percentage points, respectively. This highlights the importance of preserving exploration history and task progress during long-horizon interactions. Procedures provide reusable action guidance, particularly on BabyAI, where their removal reduces ID success by 17.3 points. Their impact is smaller on TextCraft, where ID performance is near saturation, but remains evident on OOD tasks. World models contribute most on BabyAI, where their removal reduces ID success from 64.0\% to 37.3\%. This suggests that anticipating action consequences is particularly valuable for sequential navigation and long-term planning. Finally, utility-driven retrieval improves BabyAI and TextCraft performance by filtering unreliable modules, but slightly reduces ALFWorld OOD success. Overall, these results demonstrate complementary benefits from the three modules, while highlighting the environment-dependent effectiveness in different tasks.

\begin{figure*}[!t]
    \centering
    \includegraphics[width=\textwidth]{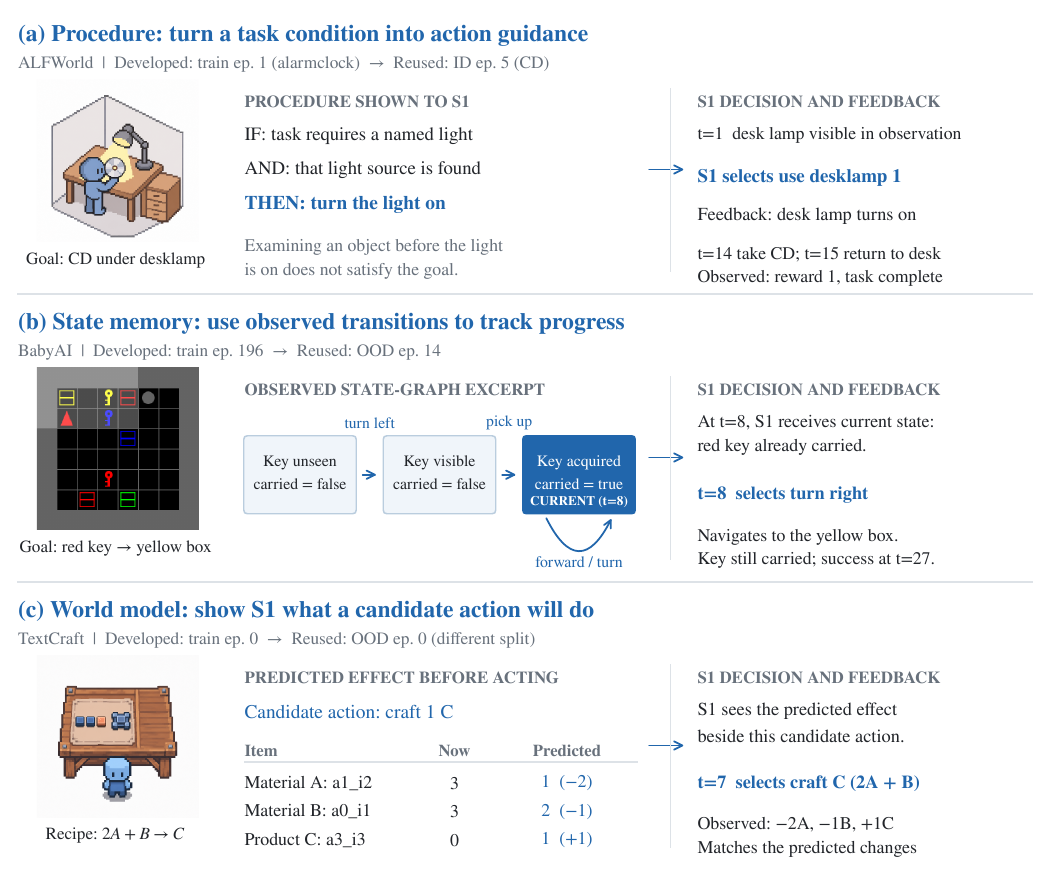}
    \caption{Three cases of harness module support in task success.}
    \label{fig:harness-case-studies}
\vspace{-5pt}
\end{figure*}

\subsection{Case Studies}
\label{sec:exp-cases}

\Cref{fig:harness-case-studies} illustrates how the three harness modules support System-1 during task execution.

\vspace{-8pt}
\paragraph{Procedure: Supplying Current-Action Prerequisites.}
In the first ALFWorld training episode, System-1 attempts to examine an alarm clock without first satisfying the required lighting condition. To address this failure, System-2 develops the \texttt{look-object-under-light} procedure, which explicitly encodes the lighting prerequisite. In ID episode 5, System-1 encounters a similar task requiring it to examine a CD under a desk lamp. The retrieved procedure provides the necessary guidance, prompting System-1 to select \texttt{use desklamp 1}. After confirming that the lamp is on, System-1 retrieves the CD and returns to the illuminated desk, completing the task in 15 steps. This case demonstrates how previously developed procedure modules transfer experience across different objects and provide actionable guidance for subsequent decisions.

\vspace{-8pt}
\paragraph{State Memory: Preserving Completed Subgoals.}
In BabyAI, repeated search cycles motivate System-2 to develop a state memory that tracks task progress. In OOD episode 14, System-1 needs to collect a red key before approaching a yellow box. The resulting state graph tracks the key's status as unseen, visible, or carried. After System-1 picks up the key at step 7, the \texttt{goalCarried} flag becomes true and remains available throughout subsequent navigation. The graph summary preserves this completed subgoal in System-1's context, allowing it to proceed toward the box without repeatedly inferring its progress from local observations. System-1 completes the task at step 27. This case demonstrates how persistent state memory preserves completed subgoals and supports robust decision-making across successive local observation changes.

\vspace{-8pt}
\paragraph{World Model: Anticipating Action Effects.}
In TextCraft, the world model \texttt{textcraft-synth-crafting}, developed in training episode 0, is reused in OOD episode 0. At step 7, the inventory contains three units each of $A$ and $B$, but no $C$. During historical interaction, the early developed world model has learned that crafting one unit of $C$ consumes two units of $A$ and one unit of $B$, leaving $A=1$, $B=2$, and $C=1$. These predicted outcomes are incorporated into System-1's context, providing it with resource requirements and expected action outcomes before execution. System-1 selects the crafting action, and the resulting inventory confirms the prediction. The crafted product then supports a subsequent step toward the target. This case demonstrates how counterfactual predictions provide System-1 with resource requirements and expected action outcomes before execution.
\vspace{-5pt}
\section{Related Work}
\label{sec:related}

\vspace{-10pt}
\noindent \paragraph{Decision Models:} Decision models enable efficient action selection through learned policies and structured prediction interfaces. Decision Transformer formulates reinforcement learning as return-conditioned sequence modeling~\citep{chen2021decisiontransformer}, while Gato employs a shared sequence model across tasks and modalities~\citep{reed2022gato}. OpenVLA extends this paradigm to robotic control by predicting actions from visual observations and language instructions~\citep{kim2024openvla}. More recently, Jev introduces a probabilistic decision framework with confidence estimates~\citep{almeida2026jev}. System1-Arena evaluates when fast decision models can replace generative reasoning across static and interactive tasks~\citep{chen2026system1arena}. Despite their efficiency, these models struggle with complex or unfamiliar tasks that require additional adequate reasoning.

\vspace{-10pt}
\noindent \paragraph{LLM Agents:} LLM agents leverage reasoning and environmental feedback to solve complex tasks. ReAct interleaves reasoning with actions~\citep{yao2022react}, while Reflexion and Voyager reuse experience through reflections and executable skills~\citep{shinn2023reflexion,wang2023voyager}. TrafficClaw extends memory-driven self-improvement to urban traffic control~\citep{lai2027trafficclaw}. USTBench evaluates LLMs' spatiotemporal reasoning in urban physical environments~\citep{lai2026ustbench}. AutoHarness synthesizes executable harnesses through environment-guided refinement~\citep{lou2026autoharness}. Adaptive Auto-Harness separates persistent development workspaces from deployable harnesses~\citep{liu2026adaptiveautoharness}. Self-evolving harness methods revise agent programs using execution feedback, supporting adaptation in long-horizon interactions~\citep{xu2026selfevolving}.

\vspace{-10pt}
\noindent \paragraph{System-1/2 Cooperative Agents:} System-1/System-2 agents combine fast execution with deliberate reasoning. SwiftSage coordinates a behavior-cloned action model and an LLM planner through a heuristic controller~\citep{lin2023swiftsage}. GR00T N1 integrates vision-language reasoning with diffusion-based action generation~\citep{nvidia2025groot}. Recent Jev-based approaches explore harness development and selective delegation. JevHarness employs an LLM to develop executable harnesses, which remain frozen during evaluation~\citep{tianyucodings2026jevharness}. JevAgentHarness delegates low-confidence decisions to an LLM and trains decision models on agent-visited states~\citep{zhu2026jevagentharness}.
\section{Conclusion}
\label{sec:conclusion}

We present JevRSI, a new System-1/System-2 collaboration framework that enables low-latency decision-making with asynchronous, LLM-driven harness recursive self-improvement. A Jev-style decision model executes System-1 fast actions, while a LLM-based System-2 develops a modular harness that integrates state memory, procedures, and world models to provide decision contexts spanning the past, present, and future. Environmental feedback recursively guides targeted harness module improvements. Experiments across three agent task datasets demonstrate favorable performance–cost trade-offs and low-latency System-1 execution. Declining development activity further suggests that continually improved harness modules reduce repeated System-2 harness development.

Our findings suggest that stronger LLM reasoning does not necessarily translate into more effective harness development. Future work will further investigate improved harness development strategies and System-1/System-2 collaboration mechanisms to simulate LLM reasoning capabilities for System-1 improvement. Moreover, the robustness of JevRSI remains untested in real-world settings (\eg embodied and autonomous driving tasks). Extending JevRSI to these environments requires richer harness representations to handle noisy observations, continuous interactions, and uncertain dynamics, while ensuring reliable decision-making.

%
%
%
%

%

\bibliography{references}

\begin{thebibliography}{29}
\providecommand{\natexlab}[1]{#1}
\providecommand{\url}[1]{\texttt{#1}}
\expandafter\ifx\csname urlstyle\endcsname\relax
  \providecommand{\doi}[1]{doi: #1}\else
  \providecommand{\doi}{doi: \begingroup \urlstyle{rm}\Url}\fi

\bibitem[Almeida(2026)]{almeida2026jev}
Almeida, D.
\newblock Introducing {System One Models} \& {Jev}.
\newblock {TypeSafe AI} Blog, September 2026.

\bibitem[Chen et~al.(2021)Chen, Lu, Rajeswaran, Lee, Grover, Laskin, Abbeel,
  Srinivas, and Mordatch]{chen2021decisiontransformer}
Chen, L., Lu, K., Rajeswaran, A., Lee, K., Grover, A., Laskin, M., Abbeel, P.,
  Srinivas, A., and Mordatch, I.
\newblock {Decision Transformer}: Reinforcement learning via sequence modeling.
\newblock \emph{arXiv preprint arXiv:2106.01345}, 2021.

\bibitem[Chen(2026)]{tianyucodings2026jevharness}
Chen, T.
\newblock {JevHarness}.
\newblock GitHub repository, 2026.

\bibitem[Chen et~al.(2026)Chen, Liu, An, Wang, and Zhou]{chen2026system1arena}
Chen, T., Liu, X., An, R., Wang, Z., and Zhou, M.
\newblock {System1-Arena}: Which decisions can be left to {System 1}?
\newblock \emph{arXiv preprint arXiv:2610.21244}, 2026.

\bibitem[Chevalier-Boisvert et~al.(2018)Chevalier-Boisvert, Bahdanau, Lahlou,
  Willems, Saharia, Nguyen, and Bengio]{chevalierboisvert2018babyai}
Chevalier-Boisvert, M., Bahdanau, D., Lahlou, S., Willems, L., Saharia, C.,
  Nguyen, T.~H., and Bengio, Y.
\newblock {BabyAI}: A platform to study the sample efficiency of grounded
  language learning.
\newblock \emph{arXiv preprint arXiv:1810.08272}, 2018.

\bibitem[Chevalier-Boisvert et~al.(2023)Chevalier-Boisvert, Dai, Towers,
  de~Lazcano, Willems, Lahlou, Pal, Castro, and
  Terry]{chevalierboisvert2023minigrid}
Chevalier-Boisvert, M., Dai, B., Towers, M., de~Lazcano, R., Willems, L.,
  Lahlou, S., Pal, S., Castro, P.~S., and Terry, J.
\newblock {Minigrid} \& {Miniworld}: Modular \& customizable reinforcement
  learning environments for goal-oriented tasks.
\newblock \emph{arXiv preprint arXiv:2306.13831}, 2023.

\bibitem[{DeepSeek-AI}(2026)]{deepseek2026v41flash}
{DeepSeek-AI}.
\newblock {DeepSeek-V4.1-Flash}: Pushing the limits of {KV} cache compression.
\newblock Model card, 2026.

\bibitem[Gandhi(2026)]{gandhi2026textcraftsynth}
Gandhi, A.
\newblock {TextCraft-Synth}.
\newblock Hugging Face dataset, 2026.

\bibitem[Gandhi et~al.(2026)Gandhi, Chakraborty, Wang, Kumar, and
  Neubig]{gandhi2026rao}
Gandhi, A., Chakraborty, S., Wang, X., Kumar, A., and Neubig, G.
\newblock Recursive agent optimization.
\newblock \emph{arXiv preprint arXiv:2605.06639}, 2026.

\bibitem[Kim et~al.(2024)Kim, Pertsch, Karamcheti, Xiao, Balakrishna, Nair,
  Rafailov, Foster, Lam, Sanketi, Vuong, Kollar, Burchfiel, Tedrake, Sadigh,
  Levine, Liang, and Finn]{kim2024openvla}
Kim, M.~J., Pertsch, K., Karamcheti, S., Xiao, T., Balakrishna, A., Nair, S.,
  Rafailov, R., Foster, E., Lam, G., Sanketi, P., Vuong, Q., Kollar, T.,
  Burchfiel, B., Tedrake, R., Sadigh, D., Levine, S., Liang, P., and Finn, C.
\newblock {OpenVLA}: An open-source vision-language-action model.
\newblock \emph{arXiv preprint arXiv:2406.09246}, 2024.

\bibitem[Lai et~al.(2025)Lai, Xu, Zhang, Liu, and Xiong]{lai2025llmlight}
Lai, S., Xu, Z., Zhang, W., Liu, H., and Xiong, H.
\newblock {LLMLight}: Large language models as traffic signal control agents.
\newblock In \emph{Proceedings of the 31st ACM SIGKDD Conference on Knowledge
  Discovery and Data Mining}, 2025.
\newblock \doi{10.1145/3690624.3709379}.

\bibitem[Lai et~al.(2026)Lai, Ning, Yuan, Chen, and Liu]{lai2026ustbench}
Lai, S., Ning, Y., Yuan, Z., Chen, Z., and Liu, H.
\newblock {USTBench}: Benchmarking and dissecting spatiotemporal reasoning
  capabilities of {LLM}s as urban agents.
\newblock In \emph{International Conference on Learning Representations}, 2026.

\bibitem[Lai et~al.(2027)Lai, Zhang, Zhou, Han, Ning, and
  Liu]{lai2027trafficclaw}
Lai, S., Zhang, P., Zhou, Y., Han, J., Ning, Y., and Liu, H.
\newblock {TrafficClaw}: Generalizable urban traffic control via unified
  physical environment modeling.
\newblock In \emph{Proceedings of the 33rd ACM SIGKDD Conference on Knowledge
  Discovery and Data Mining}, 2027.

\bibitem[Lin et~al.(2023)Lin, Fu, Yang, Brahman, Huang, Bhagavatula,
  Ammanabrolu, Choi, and Ren]{lin2023swiftsage}
Lin, B.~Y., Fu, Y., Yang, K., Brahman, F., Huang, S., Bhagavatula, C.,
  Ammanabrolu, P., Choi, Y., and Ren, X.
\newblock {SwiftSage}: A generative agent with fast and slow thinking for
  complex interactive tasks.
\newblock \emph{arXiv preprint arXiv:2305.17390}, 2023.

\bibitem[Liu et~al.(2026)Liu, Shi, Sang, He, Lin, Wei, Wang, Dumoulin, Jin, and
  Lu]{liu2026adaptiveautoharness}
Liu, Z., Shi, Z., Sang, Y., He, B., Lin, M., Wei, T., Wang, D., Dumoulin, B.,
  Jin, W., and Lu, H.
\newblock {Adaptive Auto-Harness}: Sustained self-improvement for agentic
  system deployment on open-ended task streams.
\newblock \emph{arXiv preprint arXiv:2606.01770}, 2026.

\bibitem[Lou et~al.(2026)Lou, L{\'a}zaro-Gredilla, Dedieu, Wendelken, Lehrach,
  and Murphy]{lou2026autoharness}
Lou, X., L{\'a}zaro-Gredilla, M., Dedieu, A., Wendelken, C., Lehrach, W., and
  Murphy, K.~P.
\newblock {AutoHarness}: improving {LLM} agents by automatically synthesizing a
  code harness.
\newblock \emph{arXiv preprint arXiv:2603.03329}, 2026.

\bibitem[{NVIDIA} et~al.(2025){NVIDIA}, Bjorck, Casta{\~n}eda, Cherniadev, Da,
  Ding, Fan, Fang, Fox, Hu, Huang, Jang, Jiang, Kautz, Kundalia, Lao, Li, Lin,
  Lin, Liu, Llontop, Magne, Mandlekar, Narayan, Nasiriany, Reed, Tan, Wang,
  Wang, Wang, Wang, Xiang, Xie, Xu, Xu, Ye, Yu, Zhang, Zhang, Zhao, Zheng, and
  Zhu]{nvidia2025groot}
{NVIDIA}, Bjorck, J., Casta{\~n}eda, F., Cherniadev, N., Da, X., Ding, R., Fan,
  L.~J., Fang, Y., Fox, D., Hu, F., Huang, S., Jang, J., Jiang, Z., Kautz, J.,
  Kundalia, K., Lao, L., Li, Z., Lin, Z., Lin, K., Liu, G., Llontop, E., Magne,
  L., Mandlekar, A., Narayan, A., Nasiriany, S., Reed, S., Tan, Y.~L., Wang,
  G., Wang, Z., Wang, J., Wang, Q., Xiang, J., Xie, Y., Xu, Y., Xu, Z., Ye, S.,
  Yu, Z., Zhang, A., Zhang, H., Zhao, Y., Zheng, R., and Zhu, Y.
\newblock {GR00T N1}: An open foundation model for generalist humanoid robots.
\newblock \emph{arXiv preprint arXiv:2503.14734}, 2025.

\bibitem[{OpenAI}(2026)]{openai2026gpt61sol}
{OpenAI}.
\newblock {GPT-6.1 Sol}.
\newblock Model documentation, 2026.

\bibitem[Prasad et~al.(2023)Prasad, Koller, Hartmann, Clark, Sabharwal, Bansal,
  and Khot]{prasad2023adapt}
Prasad, A., Koller, A., Hartmann, M., Clark, P., Sabharwal, A., Bansal, M., and
  Khot, T.
\newblock {ADaPT}: As-needed decomposition and planning with language models.
\newblock \emph{arXiv preprint arXiv:2311.05772}, 2023.

\bibitem[{Qwen Team}(2026)]{qwen2026qwen38}
{Qwen Team}.
\newblock {Qwen3.8-27B}.
\newblock Model card, 2026.

\bibitem[Reed et~al.(2022)Reed, {\.Z}o{\l}na, Parisotto, G{\'o}mez~Colmenarejo,
  Novikov, Barth-Maron, Gim{\'e}nez, Sulsky, Kay, Springenberg, Eccles, Bruce,
  Razavi, Edwards, Heess, Chen, Hadsell, Vinyals, Bordbar, and
  de~Freitas]{reed2022gato}
Reed, S., {\.Z}o{\l}na, K., Parisotto, E., G{\'o}mez~Colmenarejo, S., Novikov,
  A., Barth-Maron, G., Gim{\'e}nez, M., Sulsky, Y., Kay, J., Springenberg,
  J.~T., Eccles, T., Bruce, J., Razavi, A., Edwards, A., Heess, N., Chen, Y.,
  Hadsell, R., Vinyals, O., Bordbar, M., and de~Freitas, N.
\newblock A generalist agent.
\newblock \emph{arXiv preprint arXiv:2205.06175}, 2022.

\bibitem[Shinn et~al.(2023)Shinn, Cassano, Berman, Gopinath, Narasimhan, and
  Yao]{shinn2023reflexion}
Shinn, N., Cassano, F., Berman, E., Gopinath, A., Narasimhan, K., and Yao, S.
\newblock {Reflexion}: Language agents with verbal reinforcement learning.
\newblock \emph{arXiv preprint arXiv:2303.11366}, 2023.

\bibitem[Shridhar et~al.(2020)Shridhar, Yuan, C{\^o}t{\'e}, Bisk, Trischler,
  and Hausknecht]{shridhar2020alfworld}
Shridhar, M., Yuan, X., C{\^o}t{\'e}, M.-A., Bisk, Y., Trischler, A., and
  Hausknecht, M.
\newblock {ALFWorld}: Aligning text and embodied environments for interactive
  learning.
\newblock \emph{arXiv preprint arXiv:2010.03768}, 2020.

\bibitem[{TypeSafe AI}(2026)]{typesafe2026jev114}
{TypeSafe AI}.
\newblock {Jev}: For the builders {AND} dreamers.
\newblock {TypeSafe AI} Blog, October 2026.

\bibitem[Wang et~al.(2023)Wang, Xie, Jiang, Mandlekar, Xiao, Zhu, Fan, and
  Anandkumar]{wang2023voyager}
Wang, G., Xie, Y., Jiang, Y., Mandlekar, A., Xiao, C., Zhu, Y., Fan, L., and
  Anandkumar, A.
\newblock {Voyager}: An open-ended embodied agent with large language models.
\newblock \emph{arXiv preprint arXiv:2305.16291}, 2023.

\bibitem[Xu(2026)]{xu2026selfevolving}
Xu, Q.
\newblock Self-evolving harness on multiple tasks with the agent as its own
  optimizer.
\newblock \emph{arXiv preprint arXiv:2609.38372}, 2026.

\bibitem[Yao et~al.(2022)Yao, Zhao, Yu, Du, Shafran, Narasimhan, and
  Cao]{yao2022react}
Yao, S., Zhao, J., Yu, D., Du, N., Shafran, I., Narasimhan, K., and Cao, Y.
\newblock {ReAct}: Synergizing reasoning and acting in language models.
\newblock \emph{arXiv preprint arXiv:2210.03629}, 2022.

\bibitem[Zhu et~al.(2026)Zhu, Feng, Han, Wang, Yang, Tang, You, and
  Ji]{zhu2026jevagentharness}
Zhu, K., Feng, T., Han, X., Wang, H., Yang, L., Tang, X., You, J., and Ji, H.
\newblock {JevAgentHarness}: Benchmarking, deploying and training decision
  models that share control with {LLM}s.
\newblock \emph{arXiv preprint arXiv:2610.21412}, 2026.

\bibitem[Zhuang et~al.(2026)Zhuang, Su, Pan, Wong, Chen, and
  Kuo]{startlux2026decision}
Zhuang, X., Su, Z., Pan, J., Wong, N., Chen, G., and Kuo, C.-W.
\newblock {StartLux-Decision}: {AutoThink} decision models with reinforcement
  learning for calibrated deliberative decisions.
\newblock \emph{arXiv preprint arXiv:2610.17309}, 2026.

\end{thebibliography}
\bibliographystyle{icml2026}

\clearpage
\appendix

\section{Dataset Details}
\label{app:additional}
\label{app:datasets}

We evaluate three text-based interaction benchmarks. Training episodes support harness development. In-distribution (ID) evaluation measures performance on new instances from familiar task distributions. Out-of-distribution (OOD) evaluation tests transfer to unfamiliar environments or task compositions. All splits are disjoint and fixed across runs.

\subsection{ALFWorld}
\label{app:alfworld}

ALFWorld provides text-based household tasks aligned with embodied environments~\citep{shridhar2020alfworld}. Agents follow natural-language goals to navigate rooms, locate objects, and perform ordered interactions. Its six task types involve placing, examining, cleaning, heating, cooling, or collecting objects. Following the official partition boundaries, we select 200 solvable training episodes with approximately balanced coverage across task types. ID evaluation contains 140 episodes in familiar scenes with different object configurations. OOD evaluation contains 134 episodes in unseen rooms and layouts. This split tests whether agents can reuse learned interaction strategies across changing environments. Episodes are capped at 30 steps, and success requires completing the task goal.

\subsection{MiniGrid}
\label{app:minigrid}

We use BabyAI instruction-following tasks in MiniGrid, accessed through a textual interface~\citep{chevalierboisvert2018babyai,chevalierboisvert2023minigrid}. Agents navigate partially observed gridworlds, identify colored objects, and execute instructions involving object manipulation and ordered subgoals. We construct a custom split of 200 training, 150 ID, and 150 OOD episodes. Each split balances five task kinds, with 40 training episodes and 30 evaluation episodes per kind. ID introduces new maps while preserving the training task mixture. OOD introduces held-out color--object combinations and instruction sequences absent from training and ID. Duplicate map and agent configurations are excluded across splits. Episodes are capped at 150 steps. We call this benchmark BabyAI in the main results.

\subsection{TextCraft}
\label{app:textcraft}

We use TextCraft with synthetic crafting recipes~\citep{prasad2023adapt,gandhi2026rao,gandhi2026textcraftsynth}. Agents inspect recipes, manage inventory, and craft intermediate products to obtain specified target items. These tasks require planning over dependencies and tracking resource consumption. We construct a custom split from the published task pool. Training contains 200 tasks with recipe depths of 2--4. ID contains 150 distinct tasks from the same depth range. OOD contains 150 tasks with depths of 5--6, requiring longer prerequisite chains and unfamiliar goals. All splits share the same items and crafting rules, isolating generalization to deeper task dependencies.

\section{Compared Models and Methods}
\label{app:baselines}

\subsection{Decision Models}

\begin{itemize}
    \item \textbf{Jev.} Jev-1.14 is TypeSafe AI's decision model~\citep{typesafe2026jev114}. It evaluates predefined choices and returns structured answers with confidence scores. In our experiments, it receives environment observations and interaction histories, then selects the candidate action with the highest confidence.

    \item \textbf{StartLux.} StartLux-Decision is an open-weight model family that assigns probabilities to predefined choices~\citep{startlux2026decision}. We evaluate its 9B and 27B variants as standalone decision models and System-1 backbones of JevRSI. Comparing these sizes reveals how model capacity affects task success and the benefits of harness improvement.
\end{itemize}

\subsection{Large Language Models}

Each LLM serves as an agent backbone and as a System-2 harness developer in JevRSI. For LLM-agent baselines, we use the same agent framework and replace the System-1 decision model with the corresponding LLM.

\begin{itemize}
    \item \textbf{GPT-6.1 Sol.} OpenAI's general-purpose model supports reasoning, coding, and tool use~\citep{openai2026gpt61sol}. We access it through an API. It generates environment actions in the LLM-agent baseline and develops harness modules from interaction evidence when serving as System-2.

    \item \textbf{DeepSeek-V4.1-Flash.} DeepSeek's open-weight mixture-of-experts model supports long-context reasoning and tool use~\citep{deepseek2026v41flash}. We evaluate its ability to solve interactive tasks directly and to translate execution feedback into reusable harness modules for decision models.

    \item \textbf{Qwen3.8-27B.} Qwen's open-weight, 27-billion-parameter dense model supports reasoning, coding, and agentic tasks~\citep{qwen2026qwen38}. It provides a smaller LLM backbone for evaluating both direct action selection and harness development. Our experiments use textual observations and feedback.
\end{itemize}

\subsection{System-1/2 Cooperative Methods}

Both baselines use DeepSeek-V4.1-Flash as the LLM and StartLux-27B as the decision model in our implementation.

\begin{itemize}
    \item \textbf{JevHarness.} An LLM develops a task-specific harness that constructs decision inputs and combines model outputs into actions~\citep{tianyucodings2026jevharness}. Development can incorporate rewards and execution trajectories. In our evaluation, the selected harness remains fixed, and the decision model handles action selection without further LLM authoring.

    \item \textbf{JevAgentHarness.} This method shares control between a decision model and an LLM through confidence-based delegation~\citep{zhu2026jevagentharness}. The decision model handles confident choices, while the LLM takes over uncertain decisions. Our implementation invokes the LLM whenever decision confidence falls below $0.8$.
\end{itemize}

\section{Experimental Settings}
\label{app:hyperparameters}

\subsection{Asynchronous Harness Development}

We set the confidence threshold to $\eta=0.5$ in \cref{eq:development-buffer}. Each review collects three subsequent interaction steps, with remaining evidence released upon episode completion. Four asynchronous workers process a queue of up to 64 reviews, discarding the oldest when full. Each review considers up to eight relevant modules per type, alongside executed and blocked modules. System-1 continues execution during development. All LLM requests use temperature $0$, with output limits of 8,192 tokens for diagnosis and 16,384 for module development. A reasoning budget of 1,024 tokens is enabled when supported. Failed admission checks permit one repair attempt for procedures and state-memory modules, and two for world models.

\subsection{Retrieval and Decision Context}

Retrieval selects applicable modules based on contextual relevance and execution history. At each decision step, we retrieve up to $K=3$ procedures with positive relevance. At the beginning of each episode, we select the most relevant state memory program when available. Procedures and state memory modules are excluded from retrieval after $L=3$ consecutive failed episodes until revised by System-2. System-1 receives the actions and resulting observations from the eight most recent interaction steps. Meanwhile, the LLM agent baseline uses the same retrieval mechanism and decision context structure.

\subsection{State Memory}

The generated program for maintaining state memory $\sigma_m$ is limited to 6,000 characters, with execution budgets of 50 milliseconds and 20,000 Python lines per call. Admission requires successful compilation and error-free replay of the recorded episode. The graph summary $c_t^{\mathrm{past}}$ is capped at 1,200 characters and includes up to 12 recently visited states and 12 outgoing transitions from the current state. These constraints limit runtime overhead and decision-context length without restricting the underlying stored graph.

\subsection{World Model}

The generated program for predicting one-step outcomes in world models $\omega_m$ is limited to 20,000 characters, 300 milliseconds, and 500,000 executed Python lines per call. Action-effect descriptions are capped at 160 characters. Each task stores up to 400 training transitions, while prediction memory retains 200 recent transitions. Admission evaluates approximately 150 historical transitions, with at least five per task when available, plus all current-episode transitions. Initial models require at least 50\% checkable predictions and 60\% accuracy. Revisions may reduce accuracy by at most two percentage points on identical replay data. Execution errors cause rejection. System-2 receives up to eight prediction errors for revision. Selection uses accuracy smoothed with one prior success and failure, without episode-level blocking.

\subsection{Model Deployment}
\label{app:model-deployment}

We access GPT-6.1 Sol through OpenAI APIs and Jev-1.14 through TypeSafe API. DeepSeek-V4.1-Flash, Qwen3.8-27B, StartLux-9B, and StartLux-27B are deployed using vLLM. The LLM servers provide OpenAI-compatible interfaces, while StartLux returns structured action choices and confidence scores. Both local servers support contexts of up to 128K tokens. Separate inference services allow System-1 to select actions while System-2 develops harness modules asynchronously. All model weights remain fixed throughout the experiments.

\onecolumn
\raggedbottom
\section{Prompt Templates}
\label{app:prompts}

\lstdefinestyle{jevprompt}{
    basicstyle=\fontfamily{lmtt}\fontsize{8.5}{10.5}\selectfont\color{PromptInk},
    columns=fullflexible,
    keepspaces=true,
    showstringspaces=false,
    breaklines=true,
    breakatwhitespace=false,
    breakindent=0pt,
    upquote=true,
    frame=none,
    aboveskip=0pt,
    belowskip=0pt
}
\lstdefinestyle{jevprose}{
    style=jevprompt,
    breakatwhitespace=true,
    basicstyle=\normalfont\fontsize{9.5}{12}\selectfont\color{PromptInk}
}
\lstdefinestyle{jevcompact}{style=jevprompt,emptylines=0}
\definecolor{PromptInk}{HTML}{243449}
\definecolor{PromptAccent}{HTML}{275D87}
\definecolor{PromptBorder}{HTML}{D4DEE8}
\definecolor{PromptPaper}{HTML}{F8FAFC}
\definecolor{PromptHeader}{HTML}{EAF0F6}
\NewDocumentCommand{\PromptCard}{O{jevprompt} m m m O{}}{%
    \tcbinputlisting{
        enhanced,
        breakable,
        listing only,
        listing file={#4},
        listing options={style=#1},
        title={#3\hfill{\normalfont\sffamily\scriptsize\color{PromptAccent}#2}},
        title after break={#3\hfill{\normalfont\sffamily\scriptsize\color{PromptAccent}#2 / CONTINUED}},
        fonttitle=\sffamily\bfseries\small,
        coltitle=PromptInk,
        colbacktitle=PromptHeader,
        colback=PromptPaper,
        colframe=PromptBorder,
        boxrule=0.4pt,
        borderline west={1.3pt}{0pt}{PromptAccent},
        arc=1.2mm,
        left=9pt,
        right=9pt,
        top=8pt,
        bottom=8pt,
        toptitle=6pt,
        bottomtitle=6pt,
        before skip=10pt,
        after skip=12pt,
        pad at break=6pt,
        lines before break=5,
        #5
    }%
}

We provide the prompts used for action selection and harness development. Braced placeholders denote content inserted from the current interaction, module library, or admission feedback. Braces inside Python examples retain their usual meaning. The templates below follow the default configuration with world modeling enabled. Development and repair messages preserve the preceding conversation, including earlier proposals and feedback.

\subsection{System-1 Action Selection}

System-1 receives a serialized state and a typed choice question. The following layout summarizes their fields. Empty optional fields are omitted. Each neutral identifier, such as \texttt{c0}, maps to a legal action. Its description includes the predicted effect when a world-model prediction is available. The decision model returns the selected choice and its confidence.

\PromptCard{INPUT SCHEMA}{Decision context and candidate actions}{sections/prompts/s1_input_fields.txt}

The choice question uses the following instruction. Environment-specific notes and an optional hint are appended when present.

\PromptCard[jevprose]{CHOICE INSTRUCTION}{System-1 decision instruction}{sections/prompts/s1_instructions.txt}

\Needspace{9\baselineskip}
\subsection{System-2 Diagnosis}

The judge determines whether an uncertain interaction warrants a new procedure or state-memory program, a revision, or no development.

\PromptCard[jevprose]{SYSTEM MESSAGE}{Diagnose the need for harness development}{sections/prompts/s2_judge_system.txt}

The user message supplies the task, trajectory, state memory, and relevant module records. The trajectory marks uncertain steps with \texttt{>>}. The \texttt{status} field reports whether execution is ongoing or has ended in success or failure. Module records include usage, empirical utility, and failure history. The \texttt{blocked} block is included only when blocked modules exist.

\PromptCard{USER MESSAGE}{Interaction evidence and module records}{sections/prompts/s2_judge_user.txt}

\Needspace{10\baselineskip}
\subsection{Procedure Development}

When the judge selects a procedure, the following user message continues the diagnosis conversation. The \texttt{target} block contains the current module for revision and is omitted for a new module. The observation and inventory are clipped to 1{,}200 and 300 characters, respectively. The legal-action field lists up to 40 candidates and reports how many were omitted.

\PromptCard{USER MESSAGE}{Develop or revise a procedure}{sections/prompts/s2_procedure_user.txt}

\Needspace{10\baselineskip}
\subsection{State-Memory Development}

State-memory development uses the same conversation and situation fields. The generated function names the current situation using observable inputs. The framework maintains the transition graph, keeping persistent memory separate from the generated program.

\PromptCard{USER MESSAGE}{Develop or revise a state-memory program}{sections/prompts/s2_state_user.txt}

\Needspace{10\baselineskip}
\subsection{World-Model Development}

World-model development starts a separate conversation without a judge turn. The framework requests development when the environment lacks a model or observed outcomes contradict its predictions. The system message defines the model's role in constructing counterfactual action context.

\PromptCard[jevprose]{SYSTEM MESSAGE}{Define the world model's role}{sections/prompts/s2_world_system.txt}

The user message includes the current model or a notice that none exists, observed prediction errors, available graph memory, and up to 40 recent execution steps. The prediction-error block contains up to eight examples. The graph block summarizes up to 30 observed transitions. The complete template specifies the program interface, checkable prediction fields, and a generic example supplied to System-2.

\PromptCard{USER MESSAGE}{Develop or revise a one-step world model}{sections/prompts/s2_world_user.txt}

\Needspace{10\baselineskip}
\subsection{Admission Feedback and Repair}

When a proposal fails an admission check, System-2 receives the following message in the same conversation. The \texttt{problem} placeholder contains the actual validation error, such as an execution failure or insufficient prediction accuracy. System-2 must return the complete replacement module. Repair attempts follow the limits in \cref{app:hyperparameters}.

\PromptCard[jevprose]{USER MESSAGE}{Repair a rejected module}{sections/prompts/s2_repair_user.txt}

\Needspace{12\baselineskip}
\subsection{LLM-Agent Action Selection}

The LLM-agent baseline replaces the decision model with an LLM while retaining the harness context. Its system message requires one legal action and a confidence score. The user-message layout below follows the runtime field order. Optional notes, state memory, hints, and history are included only when available. If no procedure is retrieved, the learned-skills field states \texttt{(none)}.

\PromptCard[jevprose]{SYSTEM MESSAGE}{Select an action with the LLM-agent baseline}{sections/prompts/llm_agent_system.txt}

\PromptCard[jevcompact]{USER MESSAGE}{LLM-agent decision context}{sections/prompts/llm_agent_user.txt}[unbreakable]

\end{document}